\documentclass[conference,letterpaper]{IEEEtran}
\IEEEoverridecommandlockouts
\usepackage{cite}
\usepackage{amsmath,amssymb,amsfonts}
\usepackage{algorithmic}
\usepackage{graphicx}
\usepackage{textcomp}
\usepackage{xcolor}
\usepackage{hyperref}
\usepackage{algorithm}
\usepackage{float}
\usepackage{subfig}
\usepackage{caption}
\usepackage{comment}
\usepackage[normalem]{ulem}
\usepackage{flushend}
\def\BibTeX{{\rm B\kern-.05em{\sc i\kern-.025em b}\kern-.08em
    T\kern-.1667em\lower.7ex\hbox{E}\kern-.125emX}}

\usepackage{fancyhdr}
\fancypagestyle{firstpage}{
  \fancyhf{}
  \fancyhead[C]{\small To be presented at the \textit{IEEE International Symposium on Circuits and Systems (ISCAS)}, Austin, TX, USA, May 28 - June 1, 2022.}
}

\begin{document}

\title{Does Video Compression Impact Tracking Accuracy?\\
}

\author{\IEEEauthorblockN{Takehiro Tanaka, Alon Harell, and Ivan V. Baji\'{c}\thanks{ {This work was supported in part by NSERC grants RGPIN-2021-02485 and RGPAS-2021-00038.}}}
\IEEEauthorblockA{\textit{School of Engineering Science, Simon Fraser University, Burnaby, BC, V5A 1S6, Canada}}
}

\maketitle

\begin{abstract}
Everyone ``knows'' that compressing a video will degrade the accuracy of object tracking. Yet, a literature search on this topic reveals that there is very little documented evidence for this presumed fact. Part of the reason is that, until recently, there were no object tracking datasets for uncompressed video, which made studying the effects of compression on tracking accuracy difficult. In this paper, using a recently published dataset that contains tracking annotations for uncompressed videos, we examined the degradation of tracking accuracy due to video compression using rigorous statistical methods. Specifically, we examined the impact of quantization parameter (QP) and motion search range (MSR) on Multiple Object Tracking Accuracy (MOTA). The results show that QP impacts MOTA at the 95\% confidence level, while there is insufficient evidence to claim that MSR impacts MOTA. Moreover, regression analysis allows us to derive a quantitative relationship between MOTA and QP for the specific tracker used in the experiments. 

\end{abstract}


\thispagestyle{firstpage}

\section{Introduction}


\label{sec:introduction}
Video compression is ubiquitous in entertainment, monitoring, law enforcement, consumer products, and many other industries. At the same time, object tracking is the cornerstone of many video analysis applications, such as event detection and recognition, visual odometry, navigation, crowd analysis, and so on. Intuitive expectation is that compression degrades tracking accuracy. Yet, there is virtually no evidence in the available literature to support this expectation. The closest available experimental evidence links compression and object detection/classification. The earliest known study on this topic~\cite{Hase_2011} shows that JPEG and Advanced Video Coding (AVC)-based compression impacts pedestrian detection in far infrared (FIR) images, especially at lower bitrates. 
In~\cite{Wagner_2012}, the impact of JPEG2000 image compression on classifier accuracy is examined. 
Dodge and Karam~\cite{dodge_karam_2016} examined the impact of image quality, including JPEG and JPEG2000 compression, on deep neural network (DNN)-based image classification, while~\cite{Hsiang_2020} examined the impact of image compression on DNN-based object detection. All three studies revealed accuracy degradation at lower bitrates. In~\cite{Kajak_2020}, the impact of High Efficiency Video Coding (HEVC)~\cite{sullivan_overview_2012} on DNN-based based object detection was examined, with similar qualitative conclusions as earlier studies. The lack of annotated tracking data on uncompressed video was highlighted as a major challenge to making further progress in this area. 

There are a number of tracking annotations for already-compressed video,\footnote{\url{https://motchallenge.net}} so one may wonder why this data cannot be used for studying the impact of further compression on tracking? The reason is that the effect of double compression is fundamentally different from that of single compression. To see this, let $X$ be the original uncompressed signal, and
    $Y_1 = X + N_1$
be the signal after the first round of compression, where 
$N_1$ is the compression (quantization) noise. Another round of compression would produce 
    $Y_2 = Y_1 + N_2 = X + N_1 + N_2$,
where $N_2$ is the noise from the second round of compression.  {Hence,} 
the net effect on $X$ is the contamination by $N_1+N_2$.  {In a simplified case where the two noises are independent}, the probability density function (pdf) of their sum is the convolution of their individual pdf's~\cite{stark_woods_2012}: $f_{N_1+N_2}=f_{N_1}*f_{N_2}$, which will, in general, be different from either $f_{N_1}$ or $f_{N_2}$. For example, if $N_1$ and $N_2$ are uniformly distributed,\footnote{A common model for quanitzation noise.} then $f_{N_1+N_2}$ will be triangular. Hence, its effect on $X$ will not be the same as that of either $N_1$ or $N_2$. 



In this paper, we make use of the recently released dataset called SFU-HW-Tracks-v1~\cite{SFU-HW-Tracks-v1}, which contains tracking annotations for a subset of HEVC Common Test Condition (CTC) uncompressed video sequences. Since this dataset is based on uncompressed video, we are able to properly study the effects of compression on object tracking performance. Section~\ref{sec:data_methods} describes data and methods, while Section~\ref{sec:analysis} presents regression analysis of the influence of compression on tracking performance. Conclusions are presented in Section~\ref{sec:conclusions}.

\section{Data and Methods}
\label{sec:data_methods}

\subsection{Dataset} 
\label{subsec:data}
As mentioned earlier, in this study we use the recently released dataset called SFU-HW-Tracks-v1~\cite{SFU-HW-Tracks-v1}. The dataset contains object tracking annotations of 13 uncompressed test video sequences. This is an extension of the SFU-HW-Objects-v1 dataset~\cite{SFU-HW-Objects-v1}, which is currently used for exploration experiments in MPEG Video Coding for Machines (VCM).

\subsection{Methods}  \label{subsec:methods}
\begin{figure}[t]
  \centering
  \includegraphics[width=1.0\linewidth]{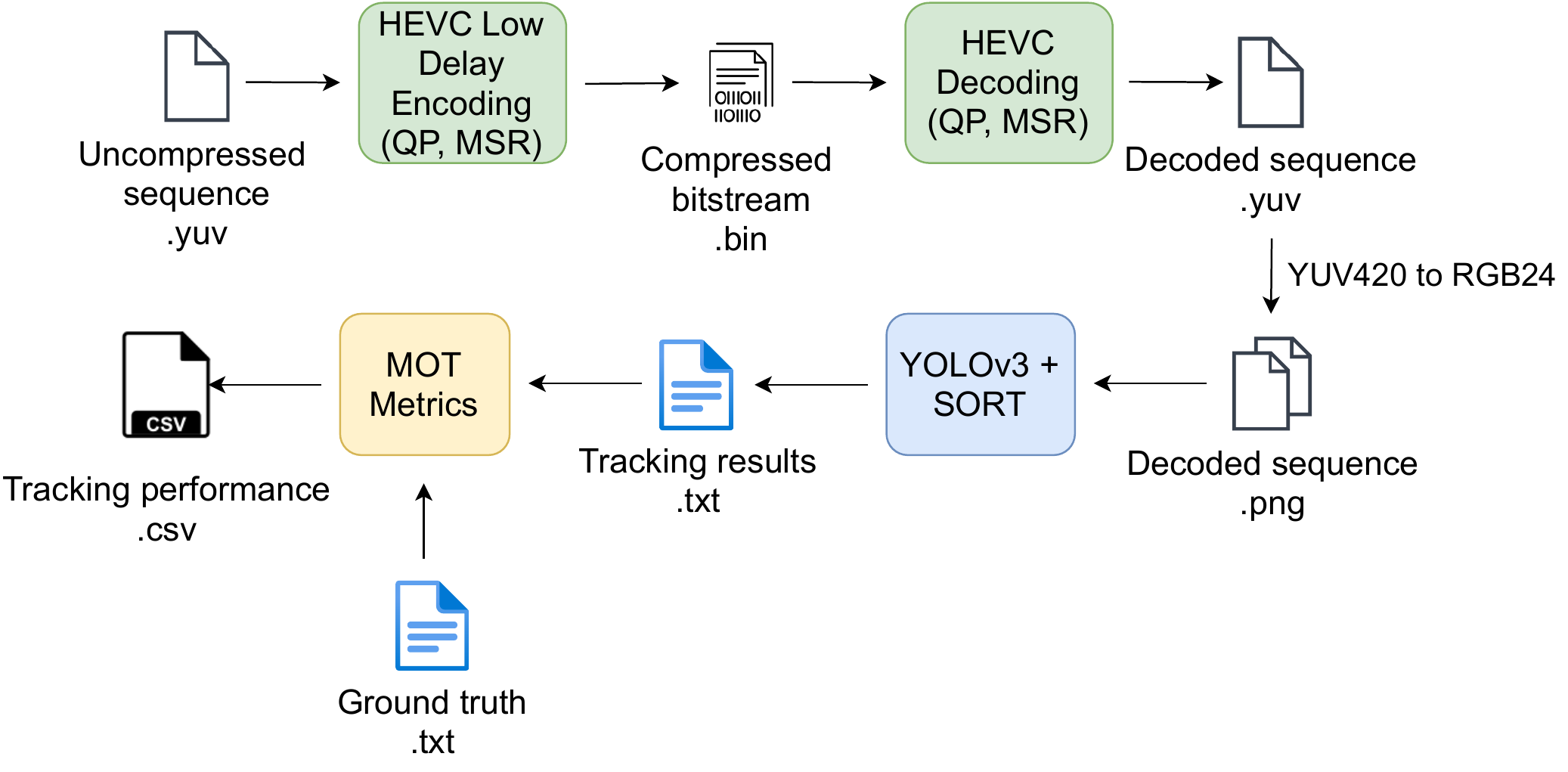}
  \caption[Diagram of experiment pipeline]
  {Experimental pipeline.}
  \label{fig:experiment_pipeline}
\end{figure}

Experimental pipeline is shown in Fig.~\ref{fig:experiment_pipeline}. An uncompressed sequence in the YUV420 format is encoded using HEVC HM version 16.20 in the Low Delay format, for various pairs of QP and motion search range (MSR), as shown in Table~\ref{tab:qp_msr_range}. The resulting bitstream is then decoded back into a YUV420 sequence, and each frame is converted to RGB format using ffmpeg and stored as PNG. Such frames are then input into the tracker (whose details are given below) and the resulting tracking output is compared to the ground truth using various Multiple Object Tracking (MOT) metrics, as described below.  

\begin{table}[t]
    \centering
    \caption[Range of control variables QP and MSR for the experiment]
    {QP and MSR values used in the experiment.}
    \begin{tabular}{|c|c|}
        \hline
        HEVC parameter & Values \\
        \hline
        \hline
        QP & [18, 22, 26, 30, 34, 38, 42, 46] \\
        \hline
        MSR & [8, 16, 32, 64] \\
        \hline
    \end{tabular}
    \label{tab:qp_msr_range}
\end{table}


\subsubsection{Tracker}
We use a tracker that follows the tracking-by-detection approach. Specifically, object detection is performed using the Ultralytics implementation\footnote{\url{https://github.com/ultralytics/yolov3},  {pre-trained on COCO~\cite{COCO}.}} of YOLOv3~\cite{redmon_yolov3_2018}, while the tracking mechanism is based on 
Simple Online Realtime Tracking (SORT)~\cite{bewley_simple_2016}. 
According to~\cite{bewley_simple_2016}, 
SORT serves as a baseline method for more sophisticated trackers, as it is constructed from well-known and interpretable components such as Kalman filtering and the Hungarian algorithm. It is worth noting that the current exploration experiments in MPEG VCM use the Joint Detection and Embedding (JDE) tracker~\cite{JDE_ECCV_2020}, whose object detection component is based on the same implementation of YOLOv3 used here. While the tracker used in our experiments is not necessarily state-of-the-art (SOTA) in terms of tracking accuracy, it is based on similar components as SOTA trackers, and provides a solid interpretable baseline for benchmarking tracking performance.

\subsubsection{Tracking metrics}
Tracking accuracy on the uncompressed and compressed videos is measured using the software framework\footnote{https://github.com/cheind/py-motmetrics} implementing various tracking metrics from~\cite{Bernadin_2008,milan_mot16_2016,Hybridboost_CVPR_2009,MTMC_ECCV_2016}. Among the various metrics available, we chose the Multiple Object Tracking Accuracy (MOTA), defined as
\begin{equation}
\text{MOTA} = 1 - \frac{\sum_{t} (\text{FN}_{t} + \text{FP}_{t} + \text{IDSW}_{t})}{\sum_{t}\text{GT}_{t}},
\label{eqn:MOTA}
\end{equation}
where $t$ is the frame index, FN$_t$ and FP$_t$ are false negatives and false positives in object detection in frame $t$, IDSW$_t$ represents the number of object ID switches in frame $t$, and GT$_t$ is the number of ground-truth trajectories in frame $t$. MOTA is the gold-standard in multiple object tracking accuracy and correlates well with human visual assessment~\cite{leal-taixe_tracking_2017}.  


\section{Analysis}
\label{sec:analysis}
\begin{figure}[t]
  \centering
  \includegraphics[width=\linewidth, trim={0.0cm 0 0 2.5cm},clip]{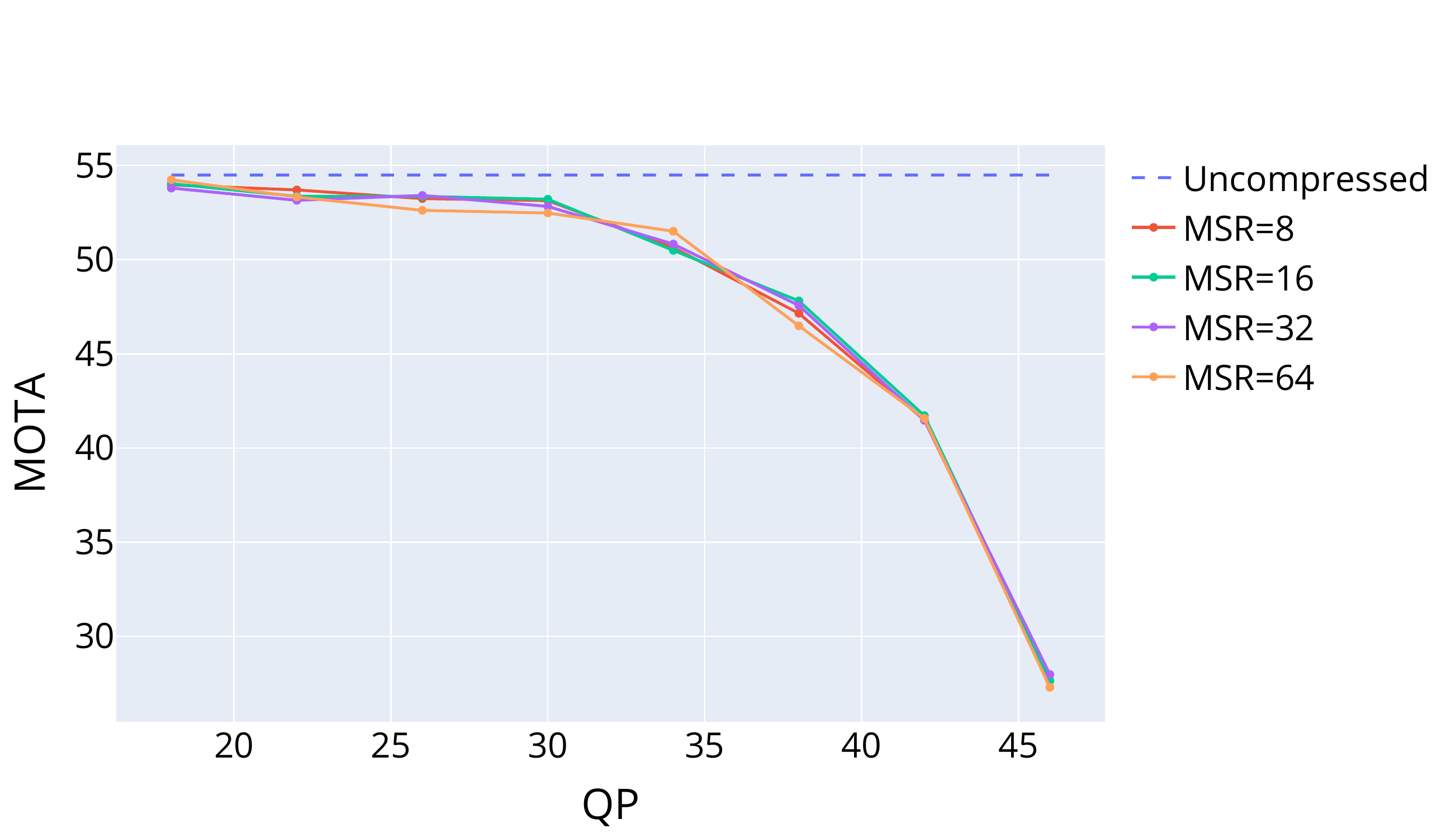}
  \caption{Average MOTA scores across all video sequences for all object classes, for uncompressed sequences (dashed line) and different (QP, MSR) pairs.}
  \label{fig:MOTA_vs_QP}
\end{figure}

The goal of our analysis is to evaluate and quantify the nature of the relationship between compression and tracking. More specifically, we are interested in the effects of the values of MSR and QP on MOTA, as explained before. As described in section~\ref{sec:data_methods}, for each pair of values of QP and MSR we have the MOTA score for multiple video sequences. We being by evaluating the average effect of QP and MSR on MOTA as can be seen in Fig.~\ref{fig:MOTA_vs_QP}. The average MOTA score is computed across all video sequences and all object classes\footnote{Here, ``all'' object classes means all classes that exist in the ground truth for each sequence.} at different QP and MSR values. As shown in the figure, MOTA scores are highest, on average, on uncompressed sequences, and degrade as the sequence is compressed, i.e., as QP increases. This seems to agree with intuition. On the other hand, MOTA scores do not seem to depend much on MSR, at least for the tracker employed in our experiments.\footnote{There are trackers, such as MV-YOLO~\cite{MV-YOLO}, that directly use motion vectors from the video bitstream, and these may be more sensitive to MSR.} In the remainder of this section we employ regression analysis~\cite{kutner_applied_2005} to statistically verify observations from Fig.~\ref{fig:MOTA_vs_QP}. The same type of analysis can be applied to other tracking metrics, other parameters (besides QP and MSR) and other trackers.

\subsection{Regression model}

Because the relationship between MOTA and QP shows clear non-linearity in Fig.~\ref{fig:MOTA_vs_QP}, we first transform QP so that MOTA and the transformed predictor variable $\text{QP}'$ form an approximately linear relationship. 
To find the appropriate transformation for QP, we experimented with various forms 
and eventually 
decided on the following one:
\begin{figure}[t]%
    \centering
    \subfloat[\centering MOTA vs. QP scatter plot]{{\includegraphics[width=3.55cm]{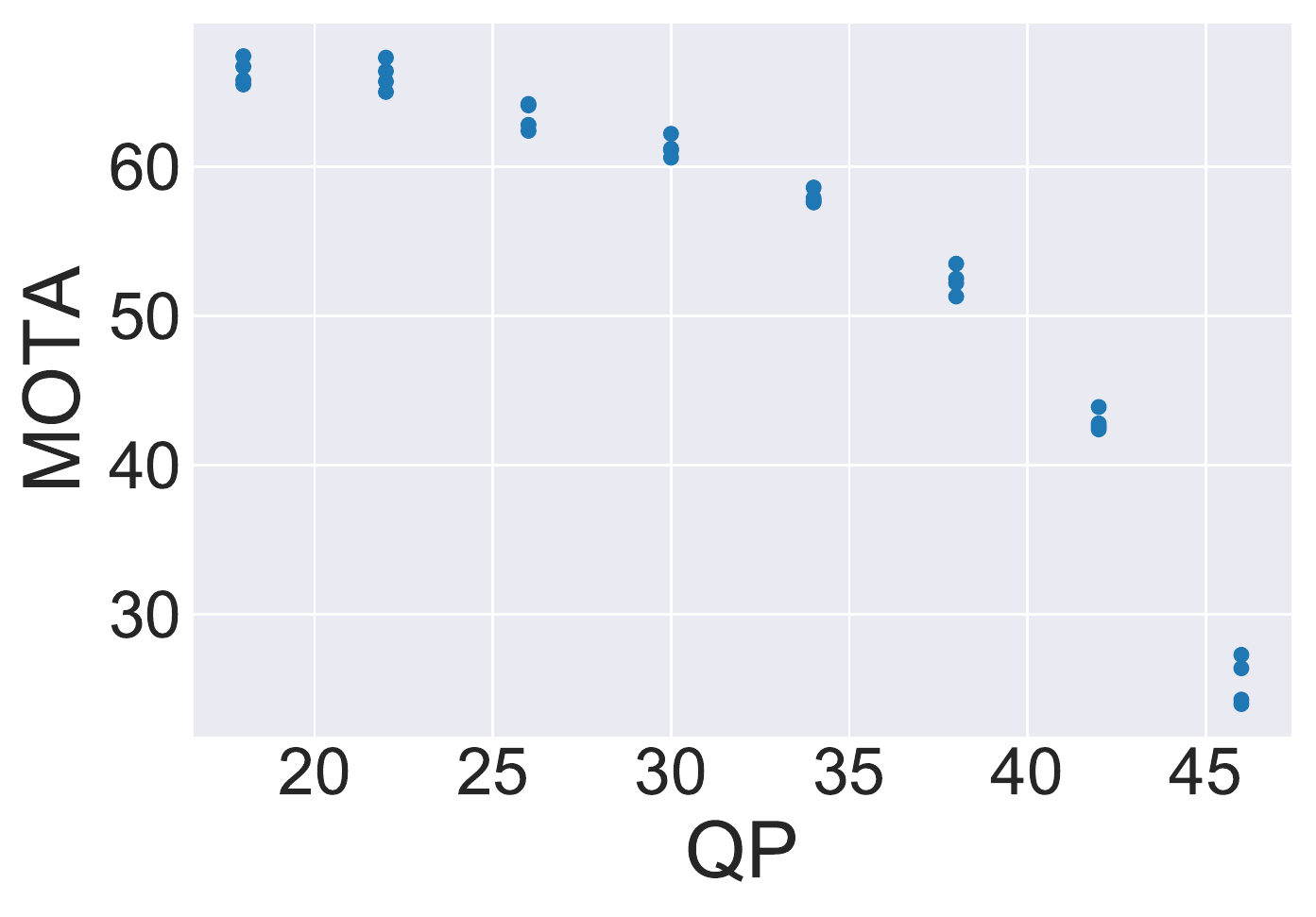} }}%
    \qquad
    \subfloat[\centering MOTA vs. $\text{QP}'$ scatter plot]{{\includegraphics[width=3.55cm]{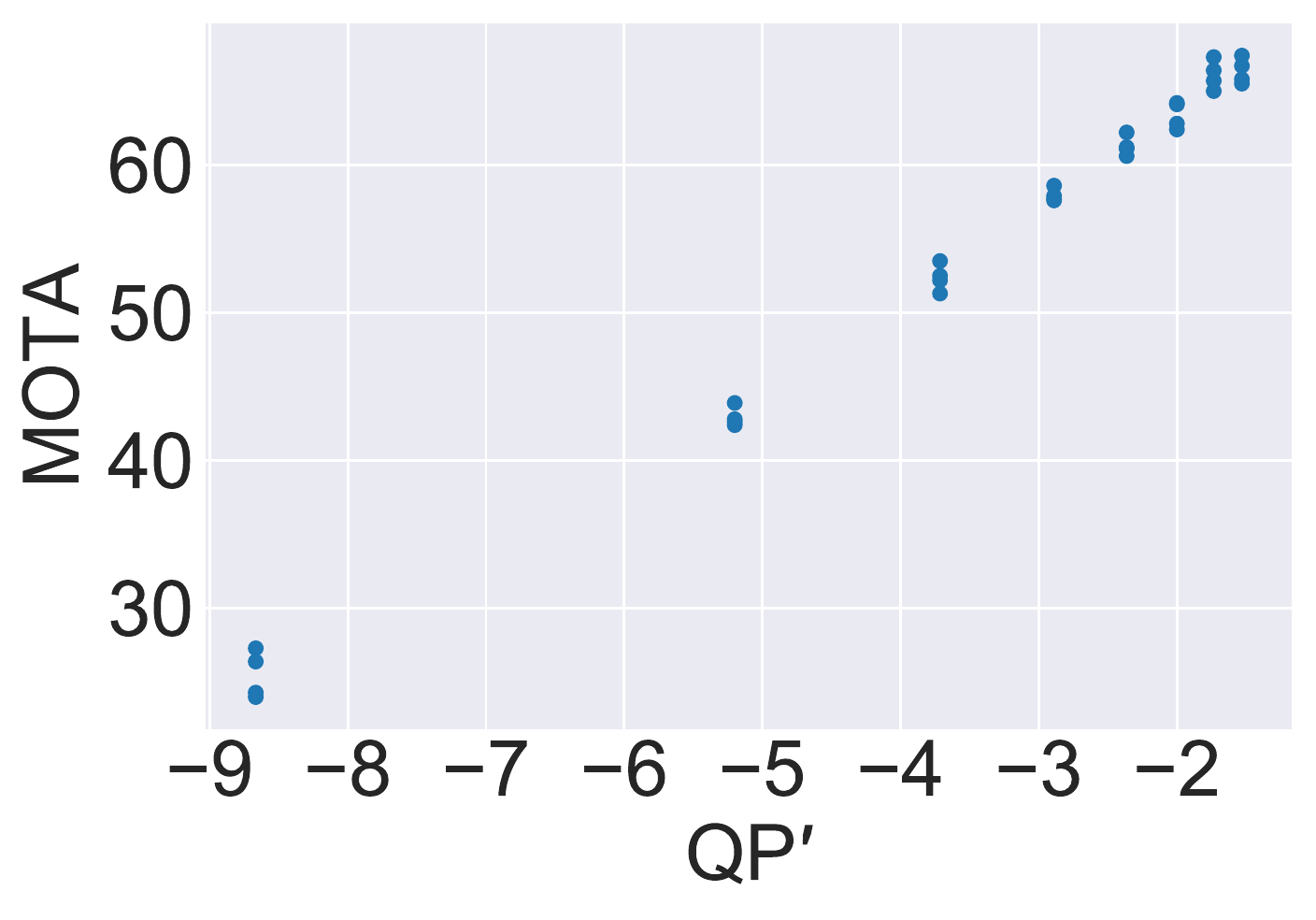} }}%
    \caption{Scatter plots of MOTA vs. QP before and after QP transformation on the BasketbalPass sequence; the four MOTA values at each QP are produced by different MSR values. }%
    \label{fig:qp_transformation}%
\end{figure}
\begin{equation}
\text{QP}' =  \frac{1}{\frac{\text{QP}}{52}-1},
\label{eqn:QP_transformation}
\end{equation}
 where $52$ was chosen to allow the full range of QP values $[0,51]$~\cite{sullivan_overview_2012}, without the danger of a zero denominator. 

The scatter plots of MOTA vs. QP and MOTA vs. $\text{QP}'$ on BasketballPass are shown in Fig.~\ref{fig:qp_transformation} where we can see the transformation indeed makes the relationship between $\text{QP}'$ and MOTA approximately linear. On the other hand, from Fig.~\ref{fig:MOTA_vs_QP} (as well as Fig.~\ref{fig:qp_transformation}), MOTA is approximately constant with MSR,  and thus already approximately linear, and appropriate for linear regression.

\begin{figure}[t]%
    \centering
    \subfloat[\centering MOTA vs. $\text{QP}'$ scatter plot ]{{\includegraphics[width=3.55cm]{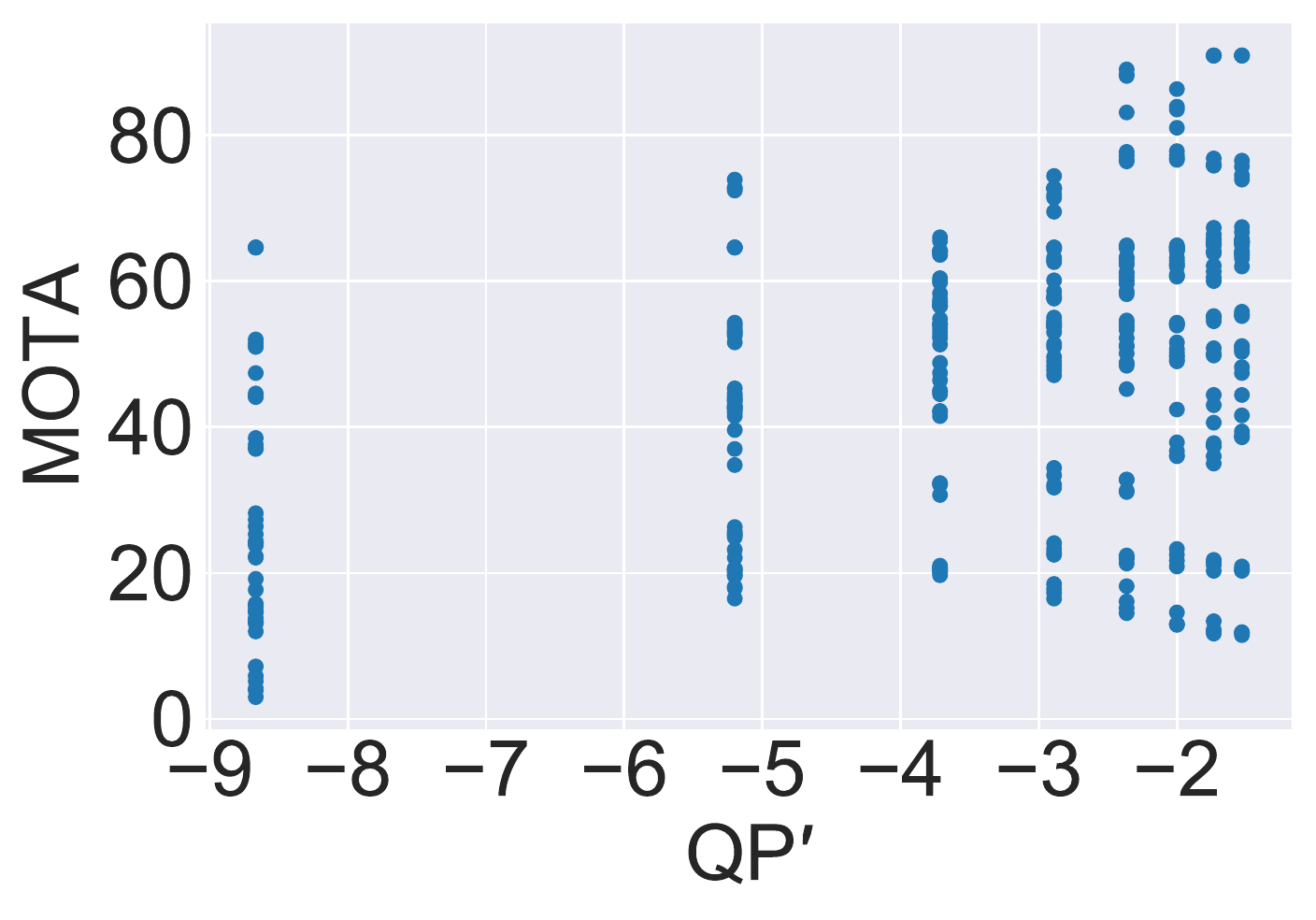} }
    \label{fig:MOTA_transformation_before}
    }%
    \qquad
    \subfloat[\centering $\text{MOTA}'$ vs. $\text{QP}'$ scatter plot]{{\includegraphics[width=3.55cm]{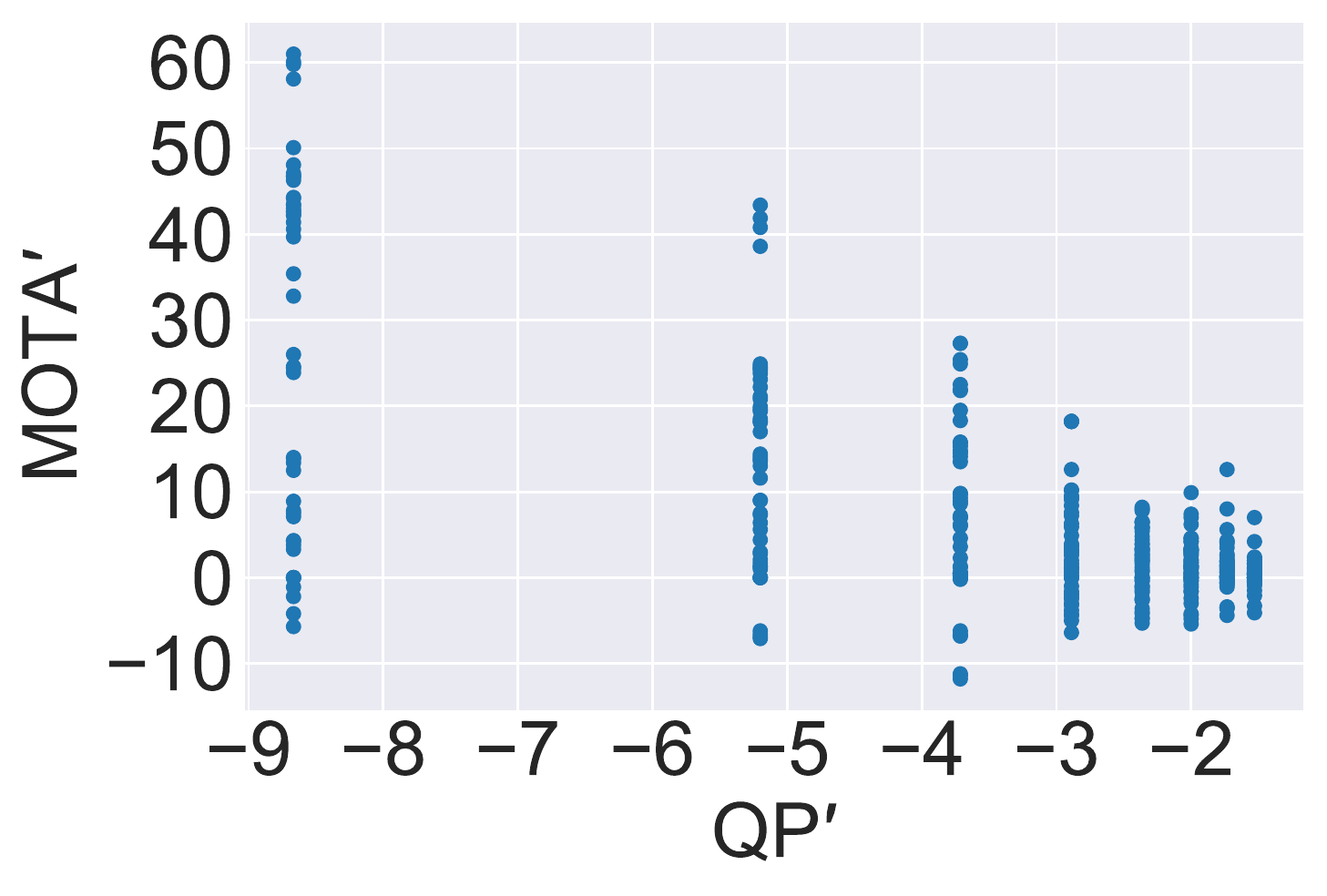} }
    \label{fig:MOTA_transformation_after}
    }%
    \caption{Scatter plots of MOTA vs. $\text{QP}'$ before and after MOTA transformation (\ref{eqn:MOTA_transformation}) across all video sequences.}%
    \label{fig:MOTA_transformation}%
\end{figure}

Because the MOTA scores, even for uncompressed videos, vary greatly between the sequences (as seen in Fig.~\ref{fig:MOTA_transformation_before}), we need to isolate the effect of compression on tracking before performing regression. 
We do this by modeling the effects of compression using a subtractive term on MOTA: 
\begin{equation}
 \text{MOTA}_i(\mathbf{Q}) = \text{MOTA}_i^0 - \text{MOTA}_i'(\mathbf{Q}),
\label{eqn:MOTA_transformation_additive}
\end{equation}
where $i$ is the sequence index, $\text{MOTA}^0$ is the $\text{MOTA}$ of an uncompressed sequence, and  $\text{MOTA}'(\mathbf{Q})$ is the  {degradation in MOTA due to compression under parameters $\mathbf{Q}$ (in our case, $\mathbf{Q}= (\text{QP}', \text{MSR})$)}. 
Re-arranging~\eqref{eqn:MOTA_transformation_additive}, we obtain:
\begin{equation}
\text{MOTA}_i'(\mathbf{Q}) = \text{MOTA}_i^0 -  \text{MOTA}_i(\mathbf{Q}),
\label{eqn:MOTA_transformation}
\end{equation}
Fig.~\ref{fig:MOTA_transformation} shows the scatter plots of MOTA and $\text{MOTA}'$ at each $\text{QP}'$, i.e., before and after MOTA transformation. We will demonstrate below that the deviations of $\text{MOTA}'$ from the regression model are approximately Gaussian. This will validate the use of statistical tests such as t-test, which rely on the Gaussianity (normality) of regression errors~\cite{kutner_applied_2005}. 

 {With QP and MOTA transformed, we can now represent a multiple linear regression model as: 
\begin{equation}
\text{MOTA}'_{i,j} = \beta_0 + \beta_1 \cdot \text{QP}'_j + \beta_2 \cdot \text{MSR}_j +\beta_3 \cdot \text{QP}'_j \cdot \text{MSR}_j + \epsilon_{i,j}
\label{eqn:regression_model}
\end{equation}
where $j$ is the index of the compression parameter combination, 
$\beta_0$, $\beta_1$, $\beta_2$, and $\beta_3$ are regression parameters, and $\epsilon_{i,j}$ is assumed to be an independent normally distributed (Gaussian) error with mean $0$ and unknown variance $\sigma^2_j$ (note that the variance only changes with $j$, not $i$). The normality of errors will be verified below.}
\begin{table}[t]
    \centering
    \caption[Regression analysis result of the MOTA score for "all" object classes across all video sequences]
    {Regression model parameter estimates across all object classes and all video sequences.}
    \resizebox{0.66\linewidth}{!}{
\begin{tabular}{|c|c|c|}
\hline
Parameter  & Estimated value  & p-value\\
\hline
\hline
$\beta_0$  & $-4.32$ & $< 10^{-6}$\\
\hline
$\beta_1$  & $-2.97$ & $< 10^{-9}$\\
\hline
$\beta_2$  & $< 10^{-2}$ &  $0.83$\\
\hline
$\beta_3$  & $< 10^{-2}$ &  $0.77$\\
\hline
\end{tabular}
    }
    \label{tab:regression}
\end{table}

 {
Allowing $\sigma^2_j$ to vary with $j$, we apply a weighted least squares method~\cite{kutner_applied_2005} to estimate $\beta$'s, with the weight 
$w(\mathbf{Q}_j)$
given by
\begin{equation}
    w(\mathbf{Q}_j) = \frac{1}{s^2(\mathbf{Q}_j)},
    \label{eqn:estimated_weight}
\end{equation}
where $s^2(\mathbf{Q}_j)$ is the sample variance of $\text{MOTA}'_{i,j}$ at a given set of parameters $\mathbf{Q}_j$ (taken over the different video sequences)}. 

\subsection{Results}
Utilizing Statsmodels~\cite{seabold_statsmodels_2010}, we obtain regression results shown in Table~\ref{tab:regression}. As seen in the table, estimated  {regression parameters} $\beta_2$ and $\beta_3$ are very small, suggesting that MOTA$'$ is not influenced by MSR or the product of MSR and $\text{QP}'$. To test the significance of 
 {each parameter,} we performed 
{the following individual hypothesis test for each $\beta_c$ as a t-test}:
\begin{equation}
    \begin{aligned}
        H_0: \beta_c = 0 \\
        H_1: \beta_c \neq 0 \\
    \end{aligned}
\end{equation}
where $c\in\{0,1,2,3\}$. The null hypothesis  {$H_0$} is $\beta_c=0$, meaning that the corresponding predictor variable does not impact the response variable $\text{MOTA}'$. The alternative hypothesis  {$H_1$} is  $\beta_c\neq0$, meaning that the corresponding predictor variable impacts $\text{MOTA}'$. The p-value for each hypothesis test is given in Table~\ref{tab:regression}. 

The results show that $H_0$ can be rejected for $\beta_0$ and $\beta_1$ (p-values are very small), meaning that these values significantly differ from $0$. In other words, QP$'$ has a significant impact on MOTA$'$. On the other hand, p-values for $\beta_2$ and $\beta_3$ are fairly large. Since they are larger than $0.05$, this means there is insufficient evidence to reject $H_0$ for these parameters at the 95\% confidence level. 
We presented individual hypothesis tests for simplicity, but a joint hypothesis test for $(\beta_2,\beta_3)$ gives similar results. Hence, from the available data, we cannot conclude that MSR impacts MOTA$'$.

\begin{figure}[t]
  \centering
  \includegraphics[width=0.8\linewidth]{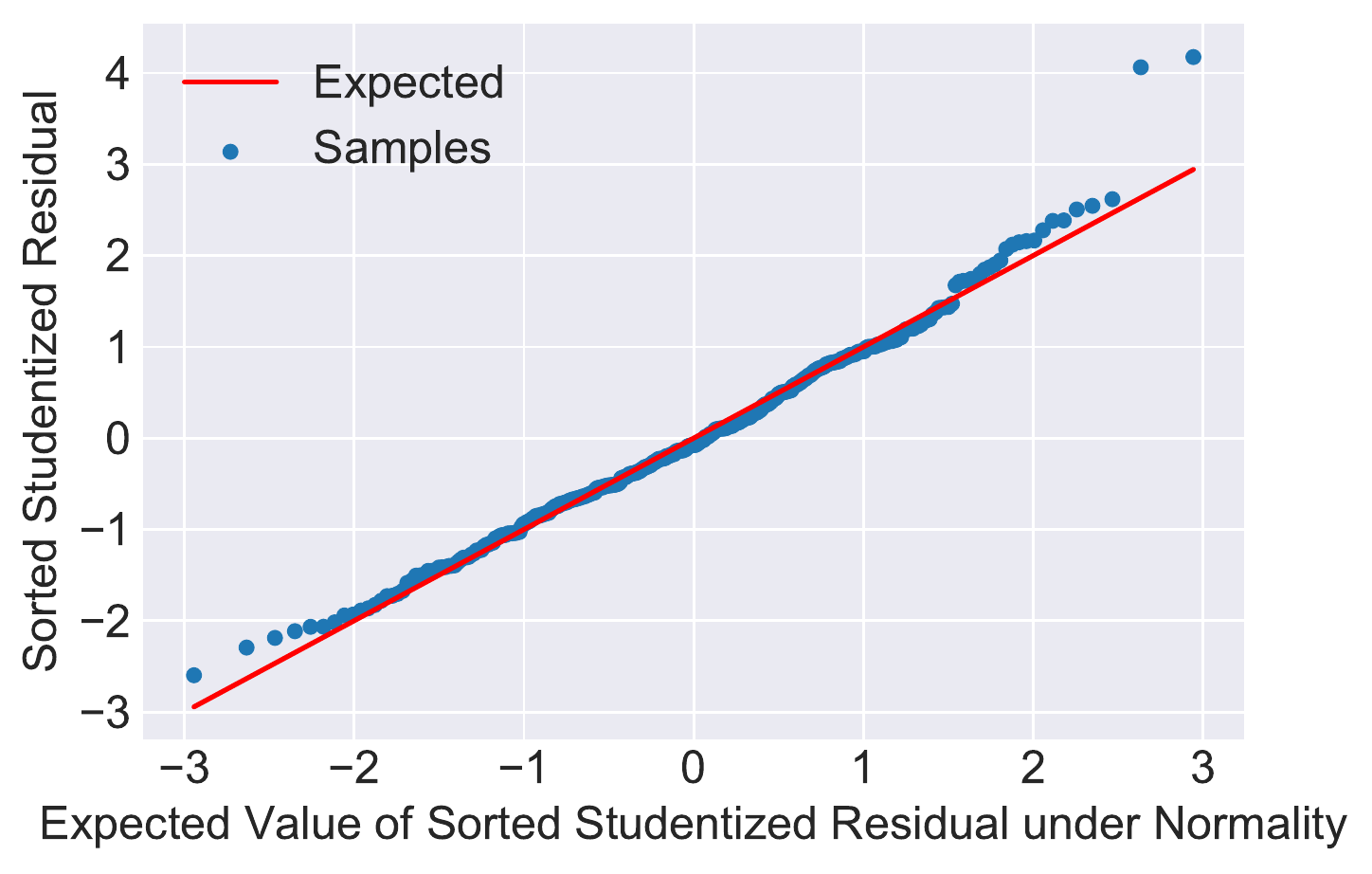}
  \caption[Testing adequacy of the regression model with normal probability plot]
  {Approximate normality of regression errors.}
  \label{fig:normal_probability_plot}
\end{figure}

  To validate our assumption regarding the normality of regression errors,  we plot the sorted studentized residuals in Fig.~\ref{fig:normal_probability_plot}, following
{~\cite{kutner_applied_2005}}.
The studentized residual $r_{i,j}$ can be represented as 


\begin{equation}
    r_{i,j} = \frac{ \text{MOTA}'_{i,j} - \widehat{\text{MOTA}'}(\mathbf{Q}_j)}{s(\mathbf{Q}_j)}
    \label{eqn:residuals}
\end{equation}
where $\widehat{\text{MOTA}'}(\mathbf{Q}_j)$ is MOTA$'$ predicted by the regression model at a given $\mathbf{Q}_j$ and $s(\mathbf{Q}_j)$ is the sample standard deviation of MOTA$'$ scores at a given $\mathbf{Q}_j$. The horizontal axis represents sorted expected values of studentized residuals under normality, which can be computed as
{~\cite[p.~111]{kutner_applied_2005}}
\begin{equation}
    \mathbb{E}[r_k] = z\left( \frac{k - 0.375}{n + 0.25} \right),
    \label{eqn:expected_residuals}
\end{equation}
where 
{$z$ is a percentile function}~\cite{kutner_applied_2005}, $k$ is the index of expected values of studentized residuals under normality, from the smallest value at $k=1$ to the largest value at $k=n$, and $n$ is the total number of measurements.\footnote{$n=384$ in our case: 8 QP values $\times$ 4 MSR values $\times$ 12 sequences. 
} As shown in Fig.~\ref{fig:normal_probability_plot}, regression residuals (blue) follow closely the red line, which shows the expected residual values under normality. In other words, the residuals are approximately normal. The coefficient of determination  {is} $R^2=0.985$, indicating close fit to the red line. This validates the regression analysis presented above. 

\begin{figure}[t]
  \centering
  \includegraphics[width=0.75\linewidth]{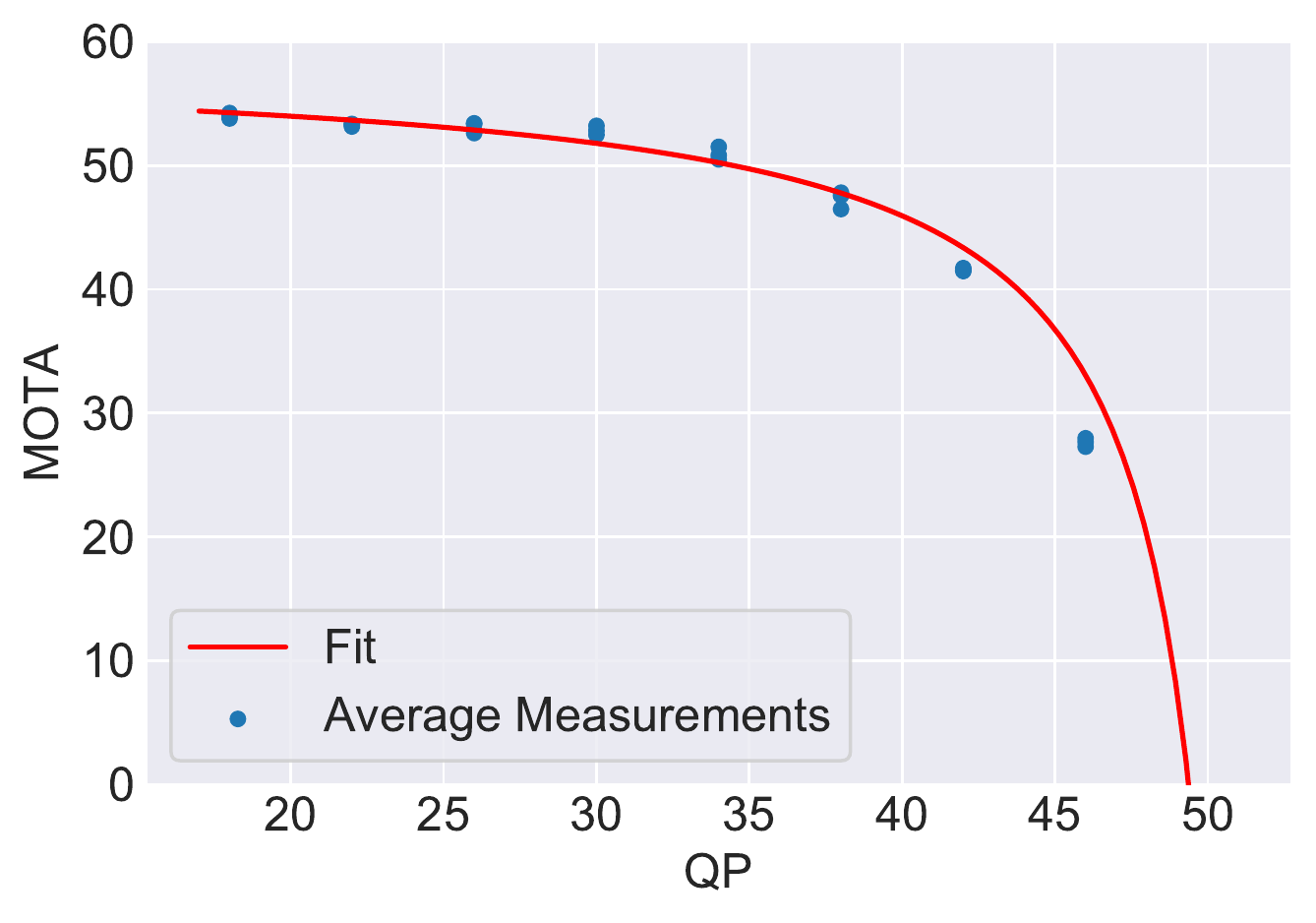}
  \caption[Average MOTA values and fitted model at different QP]
  {MOTA scores averaged over all sequences vs. QP.}
  \label{fig:MOTA_vs_QP_formulation}
\end{figure}

Using the values of $\beta_0$ and $\beta_1$ from Table~\ref{tab:regression} in~(\ref{eqn:regression_model}), and then substituting transformations from~\eqref{eqn:QP_transformation}, ~\eqref{eqn:MOTA_transformation}, we obtain the following relationship between the average MOTA across all video sequences (denoted $\overline{\text{MOTA}}$) and QP: 
\begin{equation}
    \overline{\text{MOTA}} = 58.81 + 2.97 \cdot \frac{1}{ \frac{\text{QP}}{52} - 1 }.
\label{eqn:MOTA_vs_QP_formula_substituted}
\end{equation}
Fig.~\ref{fig:MOTA_vs_QP_formulation} shows $\overline{\text{MOTA}}$ vs. QP from~(\ref{eqn:MOTA_vs_QP_formula_substituted}), as well as the average measured MOTA across all sequences at different QP values.

As seen in the figure, the agreement with measurements is fairly good. The values obtained from~(\ref{eqn:MOTA_vs_QP_formula_substituted}) for QP $>46$ should be taken with a grain of salt since there are no data points for QP $>46$, but at lower QP the agreement is good. 

\subsection{When does MOTA ``drop''?}
Using the available data, we can also answer the following question: \textit{At which QP value does MOTA start to significantly deviate from uncompressed performance?}
To answer this question, we look at the difference between uncompressed MOTA and compressed MOTA, in other words, MOTA$'$ defined in~(\ref{eqn:MOTA_transformation}). We want to determine when the average MOTA$'$ becomes significantly larger than $0$, so we formulate the following hypothesis test at each QP (or, equivalently, each QP$'$): 
\begin{equation}
    \begin{aligned}
        H_0 : \mu_{\text{MOTA}'} \leq 0 \\
        H_1 : \mu_{\text{MOTA}'} > 0 \\
    \end{aligned}
    \label{eq:mu_mota_prime}
\end{equation}
where $\mu_{\text{MOTA}'}$ is the average  $\text{MOTA}'$ at a given QP across all sequences. Note that we formulate the null hypothesis  {$H_0$} as $\mu_{\text{MOTA}'} \leq 0$ instead of $\mu_{\text{MOTA}'} = 0$, because a one-tailed t-test is more powerful than a two-tailed t-test when testing for the difference in one direction. 
\begin{table}[t]
    \centering
    \caption[t-test analysis result for MOTA]
    {One-tailed t-test results for~(\ref{eq:mu_mota_prime}).}
    \resizebox{0.43\linewidth}{!}{
\begin{tabular}{|c|c|}
\hline
{QP Value}              &               p-value \\
\hline
\hline
18 &                0.03 \\
\hline
22 &                ${< 10^{-2}}$ \\
\hline
26 &                ${< 10^{-3}}$ \\
\hline
30 &                ${< 10^{-3}}$ \\
\hline
34 &                ${< 10^{-4}}$ \\
\hline
38 &                ${< 10^{-5}}$ \\
\hline
42 &                ${< 10^{-8}}$ \\
\hline
46 &                ${< 10^{-11}}$ \\
\hline
\end{tabular}
    }
    \label{tab:t-test_MOTA}
\end{table}


One-tailed t-test results are shown in Table~\ref{tab:t-test_MOTA}. As seen in the table, all p-values are less than 0.05, so we can reject $H_0$ for all QP at the 95\% confidence level. This means that even at the lowest QP value of 18, MOTA has significantly departed (at the 95\% confidence level) from its uncompressed value. If we want to be more strict and ask for 99\% confidence, then the lowest QP for which we can reject $H_0$ is 22, because p-value for QP $=22$ is less than $10^{-2}$. Nevertheless, both  $\text{QP}=18$ and QP $=22$ are synonymous with fairly high quality compression, so we conclude that even  {high-quality} 
compression may impact tracking accuracy.  

\vspace{5pt}
\section{Conclusions}

\label{sec:conclusions}
In this paper, we analyzed the impact of video compression on object tracking accuracy using rigorous statistics. We made use of the recently released dataset of tracking annotations on uncompressed video sequences, and examined the behavior of Multiple Object Tracking Accuracy (MOTA) on HEVC-compressed sequences with different QP values and motion search range (MSR). Tracking was performed using a combination of YOLOv3 object detection and SORT tracking. Our regression analysis indicates that QP has a significant impact on MOTA, while there is insufficient evidence for MSR impact on MOTA. As a result of regression analysis, we derived a relationship between MOTA and QP for our chosen tracker, which can serve as a  guideline for estimating the impact of compression on tracking. Moreover, we showed that even low to moderate compression can have a statistically significant impact on tracking accuracy. While numerical values presented here are specific to the chosen tracker, similar analysis can be performed on other trackers as well. 

\bibliographystyle{IEEEbib}
\bibliography{refs}
\end{document}